\documentclass{article}

\usepackage{microtype}
\usepackage{graphicx}
\usepackage{subfigure}
\usepackage{booktabs} 
\usepackage{graphics} 
\usepackage{epsfig} 
\usepackage{mathptmx} 
\usepackage{times}
\usepackage{amsmath} 
\usepackage{amssymb}
\usepackage[compact]{titlesec}         
\titlespacing{\section}{0pt}{0pt}{0pt}
\AtBeginDocument{
  \setlength\abovedisplayskip{0pt}
  \setlength\belowdisplayskip{0pt}}

\usepackage{hyperref}

\usepackage[accepted]{icml2019}

\begin{document}

\twocolumn[
\icmltitle{Feature Losses for Adversarial Robustness}



\begin{icmlauthorlist}
\icmlauthor{Kirthi Shankar Sivamani}{to}
\end{icmlauthorlist}

\icmlaffiliation{to}{Electrical and Computer Engineering Department, Purdue University, West Lafayette, Indiana}

\icmlcorrespondingauthor{Kirthi Shankar Sivamani}{ksivaman@purdue.edu}

\icmlkeywords{Machine Learning, ICML}

\vskip 0.3in
]




\begin{abstract}
Deep learning has made tremendous advances in computer vision tasks such as image classification. However, recent studies have shown that deep learning models are vulnerable to specifically crafted adversarial inputs that are quasi-imperceptible to humans. In this work, we propose a novel approach to defending adversarial attacks. We employ an input processing technique based on denoising autoencoders as a defense. It has been shown that the input perturbations grow and accumulate as noise in feature maps while propagating through a convolutional neural network (CNN). We exploit the noisy feature maps by using an additional sub-network to extract image feature maps and train an auto-encoder on perceptual losses of these feature maps. This technique achieves close to state of the art results on defending MNIST and CIFAR10 datasets, but more importantly, shows a new way of employing a defense that cannot be trivially trained end to end by the attacker. Empirical results demonstrate the effectiveness of this approach on the MNIST and CIFAR10 datasets on simple as well as iterative $L_{P}$ attacks. Our method can be applied as a preprocessing technique to any off the shelf CNN. 
\end{abstract}

\section{Introduction}
\label{problem_setting_and_motivation}
    
Deep learning has achieved tremendous accuracy in solving difficult problems such as image classification \cite{resnet}, object detection \cite{yolo}, natural language processing \cite{bert},  domain adaptation \cite{sivamani2019unsupervised} and game playing \cite{alphastar}. However, recent advancements in adversarial machine learning have hindered large scale deployment of deep learning models. Szegedy et al. \yrcite{intriguingProperties} have shown that carefully crafted examples can be constructed from input images to generate incorrect outputs of high confidence. Furthermore, such inputs can be generated to specifically output a target class, and such an attack is known as a targeted attack. The \textit{adversarial} aspect of these attacks is that the changes made to the input are small enough for a human to not detect it. For image classification tasks, this is usually achieved by constraint optimization of input image pixels under an $L_{P}$ norm to only allow a maximum perturbation limit. Figures 1 and 2 show examples of adversarial images from the MNIST \cite{mnist} and CIFAR10 \cite{cifar} datasets from all 10 of their classes. 

\begin{figure}[t]
  \label{mnist_adversarial}
  \centering
  \includegraphics[width=0.48\textwidth]{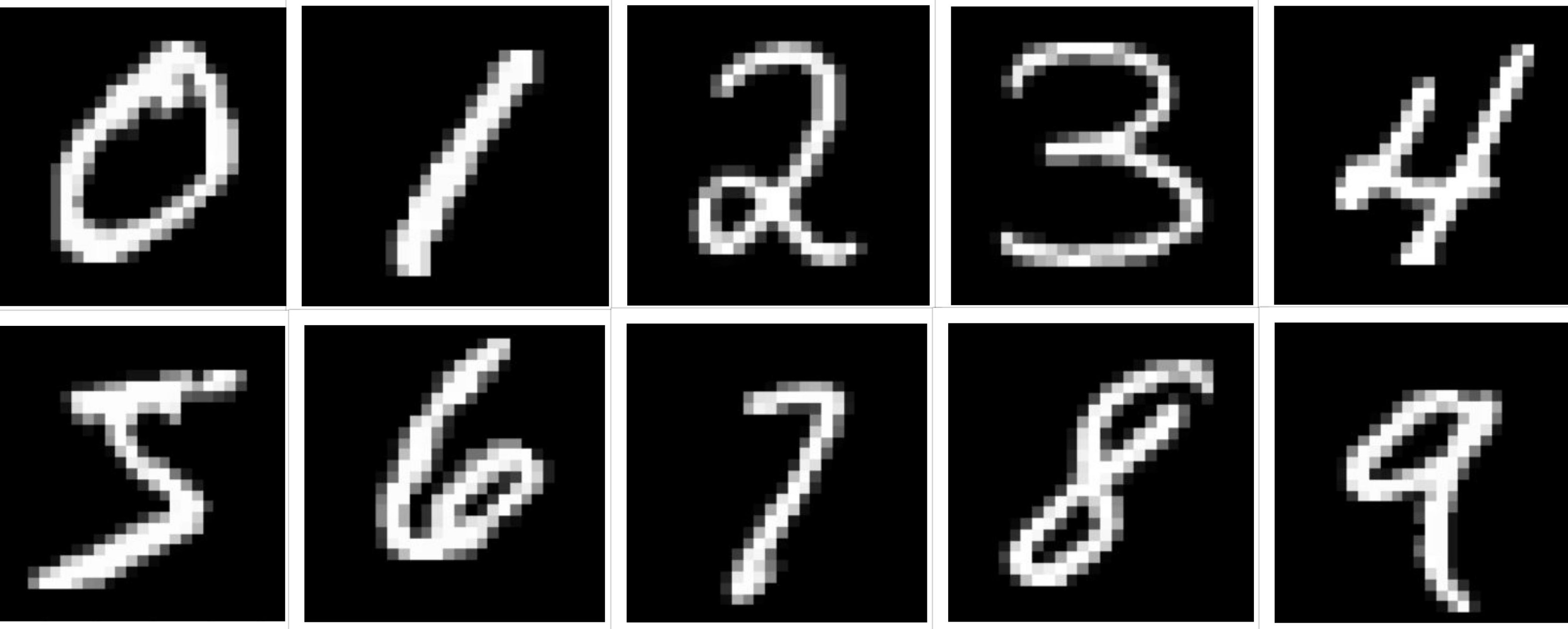}
  \caption{Sample adversarial images from MNIST. Adversarial images from all 10 classes shown are classified as a 0 with greater than 99\% confidence. (10-iteration PGD)}
\label{fig:ntd}
\end{figure}

\begin{figure}[t]
  \label{cifar_adversarial}
  \centering
  \includegraphics[width=0.48\textwidth]{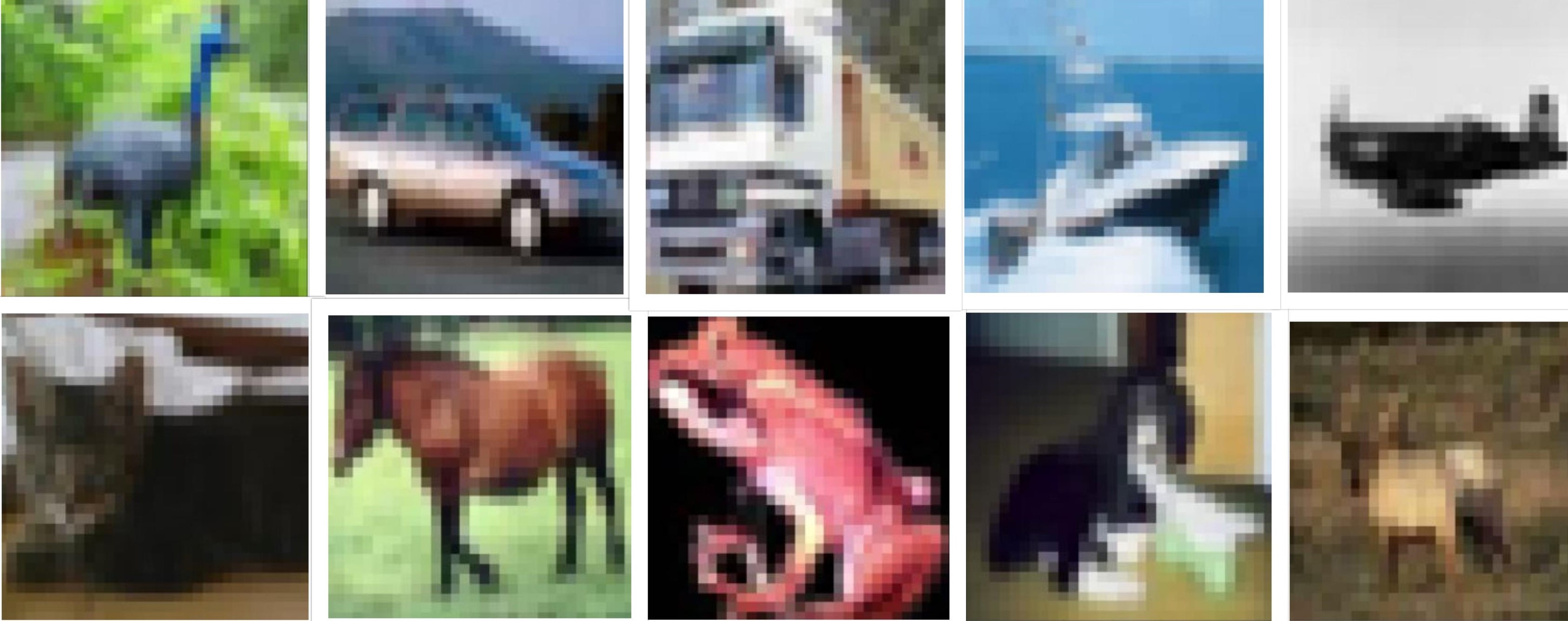}
  \caption{Sample adversarial images from CIFAR-10. Adversarial images from each class that are classified as a truck with greater than 99\% confidence. (10-iteration PGD) }
\label{fig:ntd}
\end{figure}

Several notable properties of adversarial examples have been discovered recently that make the problem worthwhile. Most surprisingly, it has been shown that adversarial examples can transfer from one network to another (source to target model) \cite{intriguingProperties}. This means the attacker does not need access to the original model to attack it. A separate model can be trained to generate adversarial samples, which can then be used as adversarial inputs to the original model. These examples get stronger (lead to highly confident incorrect predictions) as the adversary's knowledge of the target model increases. Real world examples of adversarial attacks have been explored in \cite{kurakin2016adversarial}. The authors show that adversarial images retain their properties after being printed physically, or recaptured using a camera. The latter point particularly poses threat to deep learning applications in autonomous driving, medical imaging etc. 

Current methods in adversarial defense research have two approaches: detection and classification. Carlini and Wagner \cite{adversarialNotEasilyDetected} show that detecting adversarial examples is a non-trivial task and cannot be done efficiently at the present. In this work, we aim to accurately classify adversarial samples without compromising accuracy on clean samples. Current methods to classify adversarial examples employ deep learning techniques that can be trained end-to-end. This allows the adversary to consider the defense as a part of the model, which can be attacked in the same way as the original model, nullifying the effect of the defense. Non deep learning based techniques have yielded promising results. The adversarial perturbations are imperceptible at the input level. However, Xie et al. \cite{Xie_2019_CVPR} show that these perturbation grow when passed through a deep network and show up as noise on the feature maps. Motivated by this fact, we employ a denoising autoencoder to detect and remove these noises in feature maps. 

To train the autoencoder, we use perceptual loss functions (feature losses) \cite{Johnson2016Perceptual}, previously used in super-resolution and style transfer. These loss functions employ an additional pretrained sub-network known as the loss network. Feature maps from various intermediate layers from this network are extracted and compared pixel by pixel for the input and reconstructed output of the autoencoder. This not only ensures that the final image is clean, but that it generates clean feature maps while passing through the classification network. The loss sub-network does not have to be trained for the same classification purpose. In our experiments with the MNIST and CIFAR datasets, we use the same loss network of VGG-16 \cite{vgg} pretrained on the ImageNet dataset. We argue that incorporating this additional sub-network makes it very difficult to generate adversarial examples by training it end to end.

The main contributions of the paper are summarized below:
\begin{itemize}
\item We introduce a novel method to train denoising autoencoders.
\item We create a defense strategy that cannot be trivially broken by black box attacks by training another network end-to-end.
\item We achieve 38.5\% accuracy on CIFAR10 and 83.0\% MNIST using a powerful attack.
\end{itemize}

The rest of the paper is organized as follows: Section \ref{Background} presents relevant background and related works. Our approach is presented in Section \ref{Method} followed by extensive evaluation in Section \ref{Eval}. We conclude in Section \ref{Conclusion}.

\section{Background}
\label{Background}

This work builds off of previous work in the field of adversarial defenses and machine learning. This section provides an overview of the existing related literature and the important techniques used in this study.

\subsection{Adversarial training}

Adversarial training is a defense where the network is trained on adversarial samples along with normal samples to achieve adversarial robustness \cite{fgsm}. It is the only defense that is universally accepted and guarantees improvement in accuracy. As a result, it is used as a strong baseline for many adversarial defenses \cite{Xie_2019_CVPR}. Adverserial training also addresses the trade-off in accuracy for clean and adversarial inputs for small datasets \cite{mnist, cifar} by improving accuracy on clean images, however, these results are absent in larger datasets such as ImageNet \cite{imagenet}. A major challenge posed by adversarial training is that generating adversarial examples is computationally expensive. This results in usage of the single-shot fast gradient sign method to dynamically generate adversarial images while training, which is significantly faster. This leads to sub-par adversarial training and a heavy dependence on the attack.

\subsection{Adversarial defenses}
There is a popular class of adversarial defenses that rely on minimizing the difference in a chosen metric calculated for clean and adversarial samples. Adversarial logit pairing \cite{adversarialLogitPairing} aims to minimize the mean-squared distance between the logits of the classification network. Xu et al. \cite{squeeze} reduce the search space available to an adversary by coalescing samples that correspond to many different feature vectors in the original space into a single sample. They compare the model's prediction on squeezed inputs of clean and adversarial samples to detect the difference. The author's in \cite{adaptiveNoiseReduction} assume that perturbation is a form of noise and use scalar quantization and smoothing spatial filters to denoise the inputs. Comparison with classification results of noised vs denoised version of input detects the adversary. While these methods achieve great results, they target specific defenses and have been broken trivially \cite{adversarialNotEasilyDetected}. 

\subsection{Denoising autoencoders}

Denoising autoencoders are an area of defense most relevant to this work. Several recent defensive approaches rely on input preprocessing and transformation techniques. Particularly, flavors of denoising autoencoders have yielded promising results \cite{Liao_2018_CVPR}. The neural networks mimic the process of adversarial training as a preprocessing step, instead of making the network parameters robust to adversarial inputs, a special network is trained to filter out adversarial noise such that the original classifier can process a clean image. Vanilla autoencoders suffer from the error amplification effect, in which residual adversarial noise is progressively amplified, leading to incorrect output classification. Most relevant to our approach, Liao et al. \yrcite{Liao_2018_CVPR} use high-level representation guided denoiser to overcome this problem. The high-level representation is a mean-squared loss of the output vectors of clean and adversarial images activated by the target model. They use a U-NET \cite{unet} architecture for the autoencoder.

\begin{figure}[t]
  \label{feature_noise}
  \centering
  \includegraphics[width=0.48\textwidth]{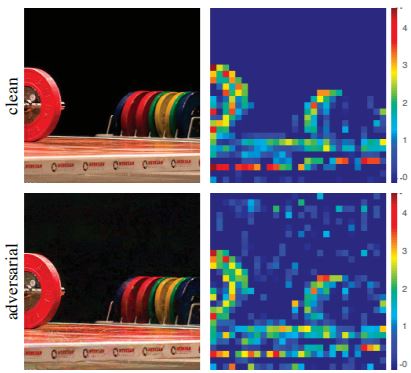}
  \caption{Feature noise. The image is adapted from Xie et al. \yrcite{Xie_2019_CVPR}. The figure shows a clean and adversarial version of an image and its corresponding feature maps. The perturbation is imperceptible in the images, but clear in the feature maps.}
\label{fig:ntd}
\end{figure}

\subsection{Feature noise}

What makes the evasion attacks adversarial is that they are quasi-imperceptible to humans. The noise injected in the training sample is small and hard to detect in preprocessing stages. However, it has been shown that \cite{Xie_2019_CVPR} these noises propagate through the network and are seen as adversarial noise on the feature maps of the target model. Figure 3 shows an example of adversarial noise present in feature maps generated using the Resnet-50 network \cite{resnet}. The noise is generally constrained by the attacker in the input images using either a matrix norm ($L_{2}$, $L_{\infty}$ etc.) or an upper bound to change per pixel ($\epsilon$). However, this constraint is missing on the feature maps of the network, resulting in the aforementioned propagation. The detection and removal of these feature noises serves as a strong motivation of adversarial defense research.  

\subsection{Perceptual and feature losses}

The idea of using a pretrained network to optimize multiple loss function at once for another network is not novel. Perceptual loss functions were first discovered by Johnson et al. \cite{Johnson2016Perceptual} for image super-resolution and style transfer. To calculate perceptual losses, a loss network pretrained for image classification is used to identify differences between content and style of an image \cite{Gatys_2016}. For neural style transfer, the pretrained loss network is used to measure differences (mean-squared error) between the feature maps of the content image and the style image. The motivation behind these perceptual losses is to gradually separate the contents and the style of an input image using feature maps from a pretrained network. Feature losses encapsulate the same concept as perceptual losses and only differ in the use case. Convolutional kernels are extracted as feature maps of any input image from a pretrained network and are compared using standard mean-squared losses. Notably, the parameters of the loss sub-network remain constant while training the autoencoder.  

\section{Method}
\label{Method}

\subsection{Threat model}

There are four possible threat models in adversarial attacks as described by Carlini and Wagner \cite{adversarialNotEasilyDetected}:

\begin{itemize}
    \item A zero knowledge attacker (black-box attack) that generates adversarial samples on a model and is not aware of any defense in place. 
    \item A perfect knowledge attacker (white-box attack) who is aware of the model architecture and parameters and also aware of the parameters and type of defense in place. 
    \item A limited knowledge attacker (grey-box attack) is aware of the neural network architecture and parameters, but unaware of the defense in place. 
    \item A variant of the gray box attack. The attacker is aware of the defense in place but unaware of the network architecture and parameters. 
\end{itemize}

As a threat model, we consider a realistic grey-box scenario where the attacker has access to the model weights and architecture but is unaware of the defense in place. This assumption gives the benefit to the attacker, although it is unlikely that the attacker has access to the model parameters. Under this threat model, We evaluate our defense on 2 popular attacks: A one shot Fast Gradient Sign Method (FGSM) attack and a more powerful iterative Projected Gradient Descent (PGD).

\begin{figure}[t]
  \label{autoencoder}
  \centering
  \includegraphics[width=0.5\textwidth]{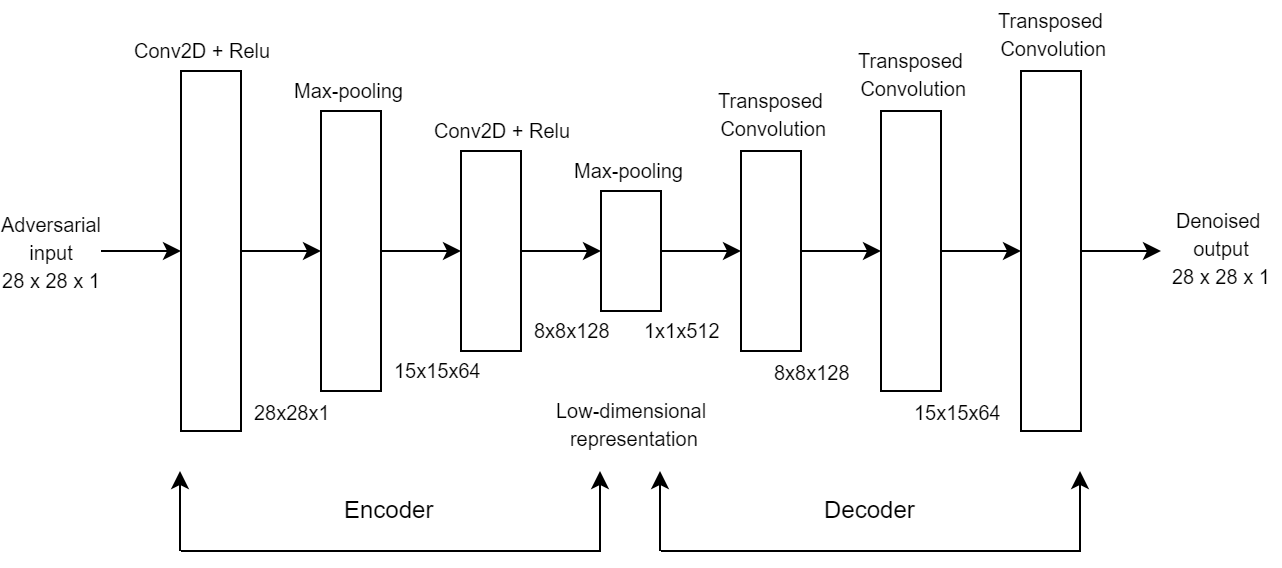}
  \caption{Autoencoder network architecture. The autoencoder consists of 7 layers in total. The encoder uses pooling to down-scale images. The decoder uses transposed convolution to upscale the images. All convolutions are activated by the rectified linear unit.}
\label{fig:ntd}
\end{figure}

\subsection{Generating adversarial samples}

Adversarial examples for both mentioned attacks are precomputed before training the autoencoder. The autoencoder is trained exclusively on adversarial samples as this is empirically shown to be most effective \cite{zhang2019theoretically}. Experiments in Section \ref{Eval} show that this does not affect accuracy on clean samples. 

\subsubsection{Fast Gradient Sign Method}

Goodfellow et al. first proposed the FGSM attack \cite{fgsm}
as a fast method to optimize the input image to convert it to an adversary. They compute gradients only once and perform a one step optimization. The main idea is to change every input pixel in the optimal direction (+ or -) upto a given perturbation limit ($\epsilon$). If x is the original input, then the perturbed image $x^{*}$ is calculated as follows:
\begin{equation} \label{eq4}
\begin{split}
    x^{*} = x + \epsilon \cdot sign(\Delta_{x}J(x, y_{true}))
\end{split}
\end{equation}

This method of using the true labels for the output y leads to the label leaking effect \cite{Kurakin2016AdversarialML} and thus we modify the equation to use the label for the target output class $y_{target}$ instead of the true labels:
\begin{equation} \label{fgsm_objective}
\begin{split}
    x^{*} = x + \epsilon \cdot sign(\Delta_{x}J(x, y_{target}))
\end{split}
\end{equation}

\begin{figure}[t]
  \label{vgg16}
  \centering
  \includegraphics[width=0.5\textwidth]{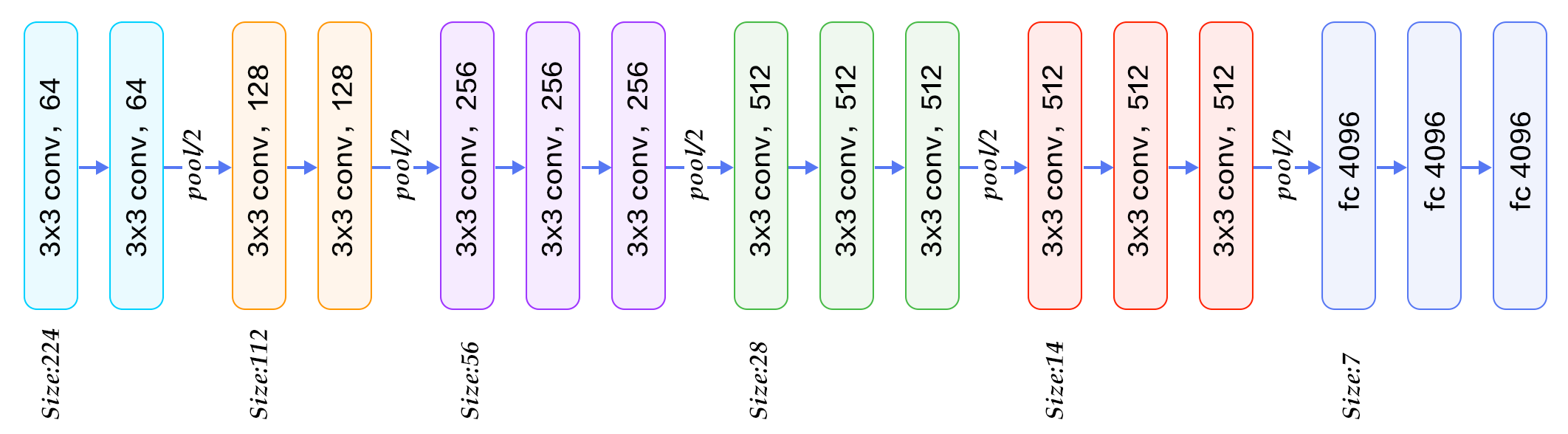}
  \caption{Feature extraction network. The VGG-16 architecture has 16 layers as seen in here. Layers 3, 7 and 10 from the back are used to extract feature maps $F_{L, 3}$, $F_{L, 7}$ and $F_{L, 10}$ of size $2\times2\times512$, $4\times4\times256$ and $8\times8\times128$ respectively. $F_{L, 3}$ corresponds to the feature map after the last red layer from the left before the max-pooling, $F_{L, 7}$ corresponds to the feature map after the second green layer from the left and $F_{L, 10}$ corresponds to the feature map after the second purple layer from the left.}
\label{fig:ntd}
\end{figure}

\subsubsection{Projected Gradient Descent}

The second attack we use is a 100 iteration version of the projected gradient descent (PGD) \cite{Madry2017TowardsDL}. Projected gradient descent is an extremely powerful first order attack as it removes any time-bound constraints on the attacker. PGD is an evolution of the iterative fast gradient sign method (IFGSM). The IGFSM attack is simply an $N$ iteration extension of the FGSM where the inputs pixels are restricted to a maximum perturbation of $\epsilon$ by clamping the input pixel space. The optimization problem is the following:

\begin{equation} \label{pgd_objective}
\begin{split}
    x^{*}_{k} = clamp_{\epsilon}(x^{*}_{k-1} + \kappa \cdot sign(\Delta_{x}J(x, y_{target}))),
\end{split}
\end{equation}

where $\kappa$ is a chosen hyper-parameter much larger or smaller than $\epsilon$. The subscript $k$ ranges from 1 to $N$ as and $x^{*}_{k}$ gets more difficult to defend as the iterations proceed. PGD starts with a random input $x_{R}$ near the original input space $x$ and performs multiple iterations of IFGSM attack to transform $x_{R}$ into an adversarial example. 

\subsection{Preprocessing network}

The preprocessing network described in this section consists of 2 main parts: The denoising autoencoder and the loss sub-network. The autoencoder comprises of an encoder and a decoder. The encoder takes in an adversarial image ($28 \times 28 \times 1$) as the input and down-scales the image into a low dimensional space ($2 \times 2 \times 8$). The decoder takes in this down-scaled decoder output and produces a clean (denoised) image of the original dimension. Figure 4 shows the architecture of the autoencoder used in our experiments.

The loss-sub network is used to generate feature maps for the loss functions that govern the training of the autoencoder. It is a pretrained image classification network using which feature maps of images can be extracted. In this work, we use a VGG-16 network pretrained on ImageNet. The feature maps of clean and adversarial samples of an input image are extracted and a linear combination of their mean-squared errors is minimized by the autoencoder, along with the image reconstruction loss. For this study, we use 3 feature maps from the VGG-16 network. $F_{L,3}$: 3 layers behind the softmax layer but before the max-pooling, $F_{L,7}$: 7 layers behind the softmax layer, and $F_{L,10}$: 10 layer behind the softmax. The resolution for these feature maps are $2\times2\times512$, $4\times4\times256$ and $8\times8\times128$ respectively. Figure 5 shows the architecture for the VGG-16 loss network that we use and the exact layers that from which the feature maps are extracted. 

Reducing the mean-squared error of multiple feature maps forces the autoencoder to denoise the input image such that the output is not only visually similar to the input, but also generates similar feature maps on a deep network to minimize the perturbations. The complete objective function for the autoencoder is given by:

\begin{equation} \label{autoencoder_objective}
\begin{split}
    L_{adv} = \alpha \cdot (F_{L,3}(x) - F_{L,3}(x^{d}))^2 \\
    + \beta \cdot (F_{L,7}(x) - F_{L,7}(x^{d}))^2 \\
    + \gamma \cdot (F_{L,10}(x) - F_{L,10}(x^{d}))^2 \\
    + \delta \cdot (x - x^{d})^2,
\end{split}
\end{equation}

where x is the input image, $x^{d}$ is the corresponding denoised adversarial sample generated, and $F_{L,n}$ is the function that produces the feature map for $x$ corresponding to the $n$ layer from the softmax in the VGG-16 network. The final term in the objective is the mean-squared image reconstruction loss. The three mean-squared errors and the reconstruction error are linearly weighted in the objective function with use of hyperparameters $\alpha$, $\beta$, $\gamma$ and $\delta$ as suggested by Johnson et al. \cite{Johnson2016Perceptual}.

\section{Evaluation}
\label{Eval}
\subsection{Datasets}
We use 2 popular datasets in the field on adversarial machine learning research. MNIST is a database of gray-scale images of digits 0-9. It consists of 60000 training images and 10000 testing images of $28 \times 28 \times 1$ resolution equally distributed among the 10 classes. CIFAR-10 is a dataset of RGB images from 10 classes of animals (e.g., frog, bird) and vehicles (e.g., aeroplane, truck). It consists of 50000 training images and 10000 testing images of $32 \times 32 \times 3$ resolution. Images from this dataset are converted to grayscale and resized to $28\times28\times1$ to match the input format for the neural networks used in this work.

\subsection{Implementation details}
\label{implementation_details}
\begin{table}[t]
\caption{Classification accuracies (\%) for FGSM and PGD attacks on the MNIST dataset.}
\label{mnist_eval}
\vskip 0.15in
\begin{center}
\begin{small}
\begin{sc}
\begin{tabular}{lcccr}
\toprule
Attack & Defense & Accuracy & Broken?\\
\midrule
FGSM    & Unsecured & 22.1 & - \\
PGD & Unsecured & 16.1 & -\\
FGSM    & Madry \yrcite{Madry2017TowardsDL} & 96.4 & $\surd$ \\
PGD & Madry \yrcite{Madry2017TowardsDL} & 92.5 & $\surd$\\
FGSM    & HGD \yrcite{Liao_2018_CVPR} & 95.9 & $\surd$ \\
PGD & HGD \yrcite{Liao_2018_CVPR} & 83.3 & $\surd$\\
FGSM    & \textbf{Ours} & 92.1 & - \\
PGD & \textbf{Ours} & 83.0 & -\\
\bottomrule
\end{tabular}
\end{sc}
\end{small}
\end{center}
\vskip -0.1in
\end{table}

We use two networks for this work. The loss network is a VGG-16 model pretrained on ImageNet for the image classification task. The architecture of the network is shown in Figure 4\footnote[2]{Training details and hyperparameters can be found on github.com/pytorch/vision/blob/master/torchvision/models/vgg.py}.

The autoencoder consists of an encoder and a decoder. The encoder takes in a $28\times28\times1$ image as the input. It consists of a $3\times3$ convolution of 3 stride and unit padding that goes from 1 to 16 channels followed the Relu activation function \cite{ReLU} and max-pooling (kernel 2 stride 2). This is followed by another $3\times3$ convolution that goes from 16 to 8 channels (stride 2 unit padding). This convolution is activated by Relu and then max-pooling (kernel 2 stride 1). The output of the encoder is $2\times2\times8$ which is fed to the decoder. The decoder upsamples this encoder output via 3 successive transposed convolutions. The first 2 transposed convolutions are followed by the Relu activation whereas the last one is followed by the inverse tangent function. The three transposed convolutions have the following configurations in order: 1) kernel size 3, stride 2, no padding, from 8 to 16 channels, 2) kernel size 5, stride 3, unit padding, 16 to 8 channels, 3) kernel size 2, stride 2, unit padding, 8 to 1 channel. 

The autoencoder was trained for 100 epochs with a batch size of 128. A learning rate of 0.001 was used and the network was optimized using the adam optimizer \cite{adam}. A weight decay of 0.00001 was used for regularizing the network. The hyperparameters $\alpha$, $\beta$, $\gamma$ and $\delta$ were chosen to be 0.00048, 0.00024, 0.00012 and 0.0013. These values are inversely related to the number of elements in the feature map. For e.g., $\alpha$ corresponds to the term containing $F_{L,3}$ in Equation \ref{autoencoder_objective}. The function $F_{L,3}$ produces a $2\times2\times512$ size feature map. Thus, $\alpha = 1/(2\cdot2\cdot512) = 0.00048$. This is done so that each term in the objective functions is weighted according to the number of elements its feature map has.

The adversarial samples generated using FGSM use an $L_{\infty}$ norm with $\epsilon =$ 0.2. Samples generated using PGD also use an $L_{\infty}$ norm with $\epsilon =$ 0.2. PGD was carried out for a 100 iterations with an input step size $\kappa$ of 0.1. Both attacks were carried out as untargeted attacks, i.e., a specific output class was not forced.

The model that we attack is a simple 4 layer convolutional neural network. There are 2 convolution:relu:max-pooling layers followed by 2 fully connected layers. The first convolution is a $5\times5$ convolution with a unit stride that goes from 1 to 20 channels. The second convolution is a $5\times5$ convolution with a unit stride that goes from 20 to 50 channels. The max-pooling layers are a size 2 stride 2 layer. The first fully connected layer takes in 800 inputs ($4\times4\times50$) and down-scales to 500 inputs. The final fully connected layer takes these 500 inputs and down-samples to give the 10 output neurons. This baseline architecture gives 99.5\% accuracy on the MNIST dataset and 92.9\% accuracy on the CIFAR10 dataset.

\subsection{Results}

We perform a total of 4 experiments as outlined in previous sections: PGD on MNIST, PGD on CIFAR-10, FGSM on MNIST, and FGSM on CIFAR10. We achieve results comparable to Madry et al. \cite{Madry2017TowardsDL} and the autoencoder approach by Liao et al. \cite{Liao_2018_CVPR}. Both of the aforementioned defenses have been broken by slightly modifying the objective functions given in Equations \ref{fgsm_objective} and \ref{pgd_objective}. The complete results for the 4 experiments have been given in Tables \ref{mnist_eval} and \ref{cifar_eval}. All experiments have been carried out using the methodology detailed in \ref{implementation_details}.

\begin{table}[t]
\caption{Classification accuracies (\%) for FGSM and PGD attacks on the CIFAR-10 dataset.}
\label{cifar_eval}
\vskip 0.15in
\begin{center}
\begin{small}
\begin{sc}
\begin{tabular}{lcccr}
\toprule
Attack & Defense & Accuracy & Broken?\\
\midrule
FGSM    & Unsecured & 22.1 & - \\
PGD & Unsecured & 16.1 & -\\
FGSM    & Madry \yrcite{Madry2017TowardsDL} & 55.5 & $\surd$ \\
PGD & Madry \yrcite{Madry2017TowardsDL} & 47.0 & $\surd$\\
FGSM    & HGD \yrcite{Liao_2018_CVPR} & 48.9 & $\surd$ \\
PGD & HGD \yrcite{Liao_2018_CVPR} & 42.2 & $\surd$\\
FGSM    & \textbf{Ours} & 43.6 & - \\
PGD & \textbf{Ours} & 38.5 & -\\
\bottomrule
\end{tabular}
\end{sc}
\end{small}
\end{center}
\vskip -0.1in
\end{table}

\section{Conclusion}
\label{Conclusion}
In this paper, we have developed an alternative approach and objective function to train denoising autoencoders as an adversarial defense. The novel objective functions are motivated by the fact that although adversarial perturbations are imperceptible at the input level, they grow and propogate forward in a feed-forward or a deep convolutional network and show up as noise in the feature maps generated by the adversarial samples on the given network. Using an additional pretrained loss network makes it non-trivial for attackers to break it using trivial backpropagation. Several design choices made in this study are open for future works and may lead to potential improvements, such as: architecture of the autoencoder, choice of the loss sub-network and its effect on accuracy, choice of feature maps within the loss sub-network etc. Evaluations are conducted on grey-box scenarios using FGSM, and a powerful 100 iteration PGD. The results achieved show promising initial results on two popular datasets MNIST and CIFAR-10, suggesting that this approach could lead to a viable long term solution in the field of adversarial defense research.

\section*{Acknowledgements}

I would like to thank Rui Wang and Praneeth Medepalli who reviewed this paper and gave valuable insights. \cite{art}

\bibliography{example_paper}
\bibliographystyle{icml2019}
\end{document}